\documentclass[letterpaper, 10 pt, conference]{ieeeconf}  
\IEEEoverridecommandlockouts                              

\usepackage[english]{babel}

\usepackage{amsthm}
\usepackage{amsmath}
\usepackage{amssymb}
\usepackage{booktabs}
\usepackage[dvips]{graphicx}
\usepackage{multirow}
\usepackage{amsfonts}
\usepackage{enumerate}
\let\labelindent\relax
\usepackage{enumitem}
\usepackage{tabularx}
\usepackage{algorithm,algorithmic}
\usepackage{bm}
\usepackage{adjustbox}
\usepackage{xcolor}
\usepackage{siunitx}
\usepackage{mdframed}
\usepackage{adjustbox}
\usepackage{authblk}
\usepackage{color}
\usepackage{xurl}
\usepackage{subcaption}
\usepackage{cite}
\makeatletter
\let\NAT@parse\undefined
\makeatother
\usepackage{relsize}
\usepackage{float}
\usepackage{pifont}
\usepackage{hyperref}
\newtheorem{definition}{Definition}
\newtheorem{theorem}{Theorem}
\newtheorem{proposition}{Proposition}
\DeclareMathOperator*{\argmin}{arg\,min}
\DeclareMathOperator*{\boundinf}{inf}
\DeclareMathOperator*{\boundsup}{sup}
\DeclareMathOperator*{\inter}{\bigcap}
\DeclareMathOperator*{\Int}{Int}

\title{\LARGE \bf
Decentralized Nonlinear Model Predictive Control for Safe Collision Avoidance in Quadrotor Teams with Limited Detection Range
}

\author{Manohari Goarin, Guanrui Li, Alessandro Saviolo, and Giuseppe Loianno
\thanks{The authors are with the New York University, Tandon School of Engineering, Brooklyn, NY 11201, USA. {\tt\footnotesize email: \{mg7363, gl1871, alessandro.saviolo, loiannog\}@nyu.edu}.}%
}
\bstctlcite{IEEEexample:BSTcontrol}

\begin{document}

\maketitle
\thispagestyle{empty}
\pagestyle{empty}

\begin{abstract}
Multi-quadrotor systems face significant challenges in decentralized control, particularly with safety and coordination under sensing and communication limitations. State-of-the-art methods leverage Control Barrier Functions (CBFs) to provide safety guarantees but often neglect actuation constraints and limited detection range. To address these gaps, we propose a novel decentralized Nonlinear Model Predictive Control (NMPC) that integrates Exponential CBFs (ECBFs) to enhance safety and optimality in multi-quadrotor systems. We provide both conservative and practical minimum bounds of the range that preserve the safety guarantees of the ECBFs. We validate our approach through extensive simulations with up to 10 quadrotors and 20 obstacles, as well as real-world experiments with 3 quadrotors. Results demonstrate the effectiveness of the proposed framework in realistic settings, highlighting its potential for reliable quadrotor teams operations.
\end{abstract}

\section*{Supplemental Material}
\textbf{Video}: \href{https://youtu.be/qTZPzcUJg0s}{https://youtu.be/qTZPzcUJg0s}

\section{Introduction}
The deployment of aerial vehicle swarms is rapidly expanding in various applications, including exploration \cite{heng2015efficient}, formation control \cite{ouyang2023formation}, search and rescue \cite{lyu2023unmanned}, and coverage \cite{cabreira2019survey}. Quadrotors are particularly well-suited for these operations due to their compact size, high maneuverability, and capability to navigate complex and dynamic environments. However, the control and coordination of a team of quadrotors introduce significant challenges, such as managing the interaction of multiple intricate dynamics, ensuring safety by avoiding inter-vehicle and environmental collisions, and being resilient to communication and sensing limitations. These challenges are further intensified as the number of robots increases, underscoring the need for decentralized control strategies that can effectively overcome centralized methods' computational and communication constraints, thereby ensuring safe, reliable, and efficient multi-quadrotor operations.

To address safety challenges, Control Barrier Functions (CBFs) have emerged as a promising framework, providing formal guarantees to maintain systems within defined safe sets, making them well-suited for safety-critical robotic applications~\cite{ferraguti2022safety}. Exponential CBFs (ECBFs) extend the applicability of CBFs to higher-order dynamics, enabling the preservation of CBF safety guarantees for complex systems such as quadrotors~\cite{khan2020barrier, xu2018safe, du2023simultaneous, jin2023safety}. These advancements align with the need for robust safety measures in multi-quadrotor systems. However, despite their theoretical promise, current implementations often rely on simplified models and neglect practical constraints, such as motor thrust limitations, which can compromise the effectiveness and reliability of these systems in real-world operations~\cite{palani2023control, mali2021incorporating, wang2024dual}.

Moreover, the implementation of CBFs and ECBFs in decentralized control systems is hindered by sensing and communication limitations, such as restricted detection ranges. These constraints undermine the safety guarantees provided by these control methods, particularly in environments where timely and accurate decision-making is critical.

\begin{figure}[t]
    \centering
    \includegraphics[width=\columnwidth]{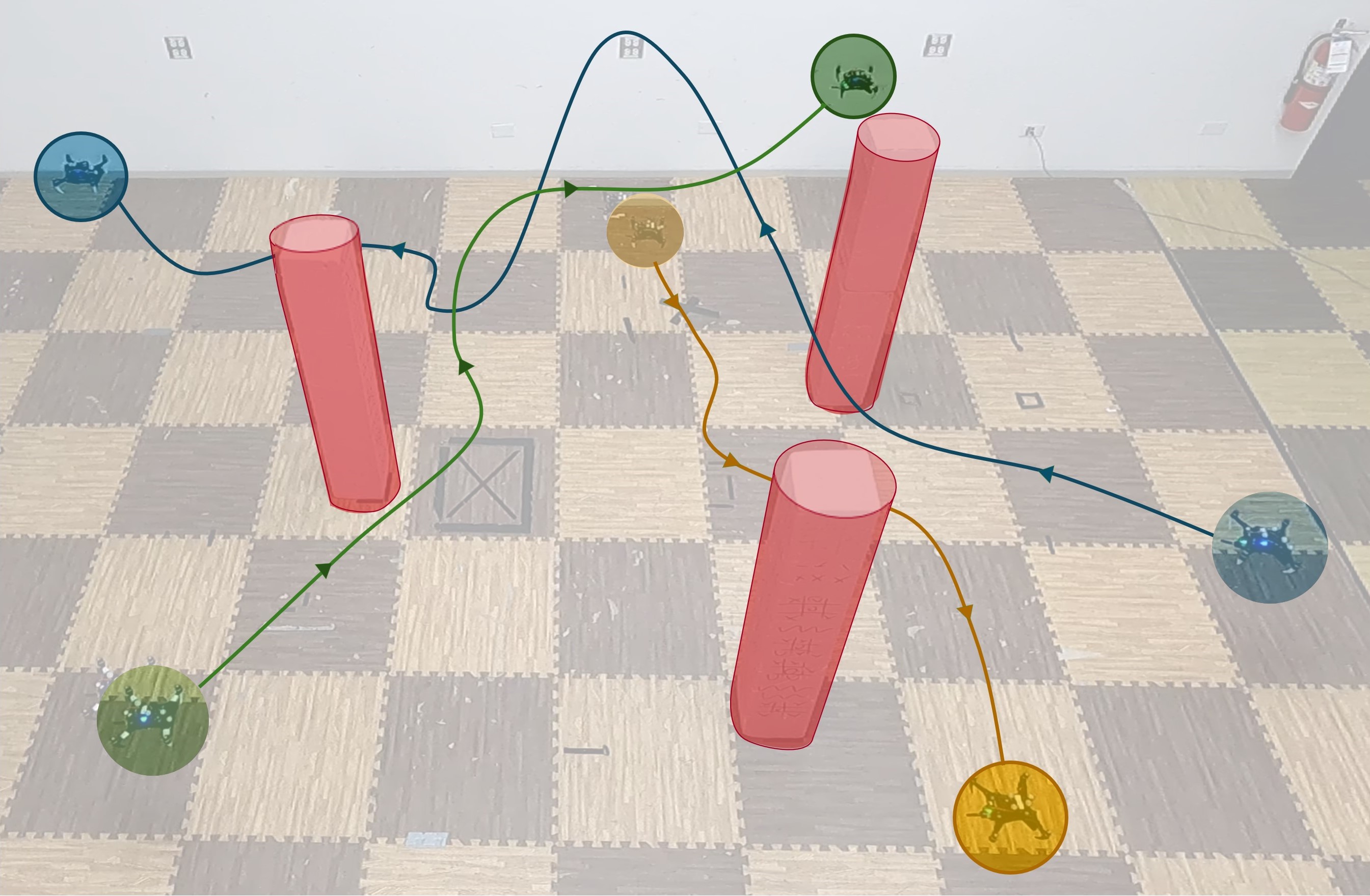}
    \caption{
    \textbf{Decentralized Safe Control for Multi-Quadrotor systems.}
    The proposed control strategy guides three quadrotors (blue, green, yellow) to safely maneuver around obstacles (red), demonstrating successful collision avoidance.
    }
    \vspace{-14pt}
    \label{fig:rw_exp}
\end{figure}

This work aims to fill the aforementioned gaps by proposing a novel decentralized Nonlinear Model Predictive Control (NMPC) that incorporates ECBFs in limited detection scenarios to enhance safety and optimality for multi-quadrotor control under motor thrust constraints.
The primary contributions of this work are the following
\begin{itemize}[left=0pt]
    \item We develop a novel decentralized NMPC approach featuring pair-wise ECBFs to safely control a team of quadrotors under motor thrust constraints and limited detection range. 
    \item We analyze the safety guarantees of the ECBFs under limited detection range. Specifically, we establish a conservative lower bound of the range to guarantee safety across all ECBFs. For practical use, we also derive a less restrictive bound tailored to worst-case quadrotor-to-quadrotor or quadrotor-to-obstacle interactions.
    \item We validate the proposed approach through extensive simulations and real-world experiments with the control framework deployed and running on-board, demonstrating the  practicality of the approach. 
\end{itemize}

\section{Related Works}
Various methods have been explored for multi-robot collision avoidance in the literature \cite{raibail2022decentralized} but CBFs are popular for providing strong safety guarantees in control systems~\cite{ferraguti2022safety}. 

\textbf{Decentralized CBF-based control methods.}
Decentralized multi-robot collision avoidance using CBFs is widely studied in the literature \cite{chen2020guaranteed,parwana2023rate, palani2023control,mali2021incorporating,wang2024dual, ibuki2020optimization,mestres2024distributed,catellani2023distributed}. Decentralization is crucial for achieving scalability in terms of number of agents and obstacles in the environment, and addressing real-world sensing and communication limitations \cite{raibail2022decentralized}. Among these methods, the works \cite{chen2020guaranteed,ibuki2020optimization,mestres2024distributed} apply position-based barrier functions and first-order CBF constraints to wheeled robots. However, traditional first-order CBFs are not valid for higher-order systems subject to actuation constraints such as quadrotors~\cite{ames2019control}. The authors in \cite{palani2023control, catellani2023distributed} simplify the problem by optimizing the displacement and velocity of quadrotors. However, reducing the model to a first-order system leads to inaccuracies and reduced safety in real world conditions. Conversely, the authors in \cite{mali2021incorporating} manually incorporate velocity information into the barrier function to create valid CBFs. Recently, higher-order CBFs (HOCBFs) \cite{xiao2021high} have been developed to handle general high-order dynamics and preserve the forward invariance property of their safe set. They have been used in multi-robot settings in works like~\cite{wang2024dual,parwana2023rate} but using double integrator dynamics. Additionally, some works also leverage the differential flatness property of quadrotors along with first and higher-order CBFs for trajectory planning \cite{wang2017safe,zhang2024cooperative,abichandani2023distributed}. Overall, these works overlook realistic actuation constraints like motor thrust limitations in quadrotors which can compromise the feasibility of both trajectories and safety constraints.

In contrast, in single quadrotors applications, nonlinear optimization-based control solutions with ECBFs~\cite{khan2020barrier,xu2018safe,du2023simultaneous,jin2023safety} are proposed, handling concurrently the full dynamics of the quadrotor, safety and actuation constraints. 
ECBFs are a specific form of higher-order CBFs that utilize linear control theory to enforce desired safety behaviors within a system. They have proven effective in challenging scenarios, and eliminate the need for manually designed barrier functions.
Moreover, integrating CBFs into MPC has become increasingly popular due to MPC's predictive capabilities, which enhances solution optimality while simultaneously enforcing safety constraints \cite{zeng2021enhancing,thirugnanam2022safety,mundheda2023control,minh2022safety,marvi2019safety,mali2021incorporating,jiang2023incorporating,liu2024safety}. In this paper, we aim to unify second-order ECBFs with a decentralized Nonlinear MPC (NMPC) for navigation of multi-quadrotor systems, combining the strengths of ECBFs and NMPC to achieve safe, optimal control under motor thrust constraints.

\textbf{CBF safety guarantees under Limited Detection Range.}
Many works restrict the operating range of robots to a limited neighborhood, thereby dropping the complexity of the optimization problem \cite{wang2017safety,palani2023control} and accounting for sensing and communication restrictions \cite{mestres2024distributed,palani2023control,ibuki2020optimization,catellani2023distributed}. 
However, these restrictions can significantly compromise the CBF safety guarantees.  While some works address the feasibility of the optimization problem or the compatibility of multiple CBFs \cite{ibuki2020optimization,wang2017safety,catellani2023distributed,parwana2023rate}, few have studied how limited sensing and communication impact the safety guarantees of the CBFs. The authors in \cite{wang2017safety} propose a neighborhood range beyond which the pair-wise CBF constraints are automatically satisfied, allowing distant robots to be ignored without impacting the robots' safety, in the case of first-order CBFs. 
In this paper, we extend this analysis to the safety guarantees of the second-order ECBFs of our control framework under limited detection (through sensing or communication) range.

\section{Preliminaries on Exponential Control Barrier Functions}
In the following, we denote column vectors with bold notation like $\mathbf{x}$, matrices with capital notation like $\mathbf{A}$, and scalars with unbold notation like $h$.
For preliminaries on CBF foundations, we refer the reader to \cite{ames2019control,ferraguti2022safety}.
Exponential Control Barrier Functions (ECBFs) \cite{ames2019control,nguyen2016exponential} were introduced to enforce safety to high relative-degree systems, i.e. when the first time-derivative of the barrier function does not depend on the control input.

Let us consider a nonlinear control affine system
\begin{equation}
    \dot{\mathbf{x}} = f(\mathbf{x})+g(\mathbf{x})\mathbf{u}.
    \label{affine_sys}
\end{equation}

We define the safe set $\mathcal{C}$ as the superlevel set of a continuously differentiable function $h:\mathcal{X}\subset \mathbb{R}^n \rightarrow \mathbb{R}$
\begin{equation}
    \mathcal{C} = \{ \mathbf{x} \in\mathcal{X}\subset \mathbb{R}^n: h(\mathbf{x}) \geq 0\},
\end{equation}
$h(\mathbf{x})$ has arbitrarily high relative degree $r\geq 1$ with respect to $\mathbf{u}$, and its $r$th time-derivative can be formulated as
\begin{equation}
    h^{(r)}(\mathbf{x},\mathbf{u}) = L_f^r h(\mathbf{x}) + L_g L_f^{r-1} h(\mathbf{x})\mathbf{u},
\end{equation}
with $L_g L_f^{r-1} h(\mathbf{x}) \neq 0$ and $L_g L_f h(\mathbf{x}) = L_g L_f^2 h(\mathbf{x}) = ... = L_g L_f^{r-2} h(\mathbf{x})=0$, $\forall \mathbf{x} \in \mathcal{X}$. $L_f h(\mathbf{x})$ and $L_g h(\mathbf{x})$ represent the Lie derivatives of $h(\mathbf{x})$ along the vector fields $f(\mathbf{x})$ and $g(\mathbf{x})$ respectively.

\begin{definition} \label{def_ecbf}
    $h$ is an exponential control barrier function if there exists $K_\alpha \in \mathbb{R}^r $ such that for the control system~\eqref{affine_sys}
    \begin{equation*}
        \sup_{\mathbf{u} \in U} \bigg\{ L_f^r h(\mathbf{x}) + L_g L_f^{r-1} h(\mathbf{x})\mathbf{u} + K_\alpha \eta_b(\mathbf{x})\bigg\} \geq 0, \forall \mathbf{x} \in \Int(\mathcal{C})
    \end{equation*}
    \text{with} $\eta_b (\mathbf{x}) = \begin{bmatrix}
            h(\mathbf{x}) & L_f h(\mathbf{x}) & \cdots & L_f^{r-1} h(\mathbf{x})
        \end{bmatrix}^\top$ and $K_\alpha = \begin{bmatrix}
            \alpha_1 & \cdots & \alpha_r
        \end{bmatrix}$.
\end{definition}
In order to enforce the forward invariance of $\mathcal{C}$, defined in \cite{ames2019control}, the tunable gain row vector $K_\alpha$ needs to satisfy certain properties, summarized in the following.

We define the set of functions $\nu_i:\mathcal{X}\rightarrow \mathbb{R}$ and the corresponding superlevel sets $\mathcal{C}_i$ as follows
\begin{subequations}
\begin{align*} 
    &\nu_0 (\mathbf{x}) = h(\mathbf{x}),  \hspace{73.5pt} \mathcal{C}_0 = \mathcal{C} \\
    & \nu_1(\mathbf{x}) = \dot{\nu}_0(\mathbf{x}) + p_1 \nu_0(\mathbf{x}), \hspace{25pt}\mathcal{C}_1 = \{ \mathbf{x}: \nu_1(\mathbf{x})\geq 0\} \\
    &\quad \vdots \hspace{125pt} \vdots \nonumber\\ & \nu_r(\mathbf{x}) = \dot{\nu}_{r-1}(\mathbf{x}) + p_r \nu_{r-1}(\mathbf{x}), \hspace{5pt}\mathcal{C}_r = \{ \mathbf{x}: \nu_r(\mathbf{x})\geq 0\}
\end{align*}
\end{subequations}
The coefficients $p_1, ..., p_r$ are the roots of the polynomial $\lambda^r + \alpha_r \lambda^{r-1}+ ... + \alpha_2 \lambda + \alpha_1 = 0$ \cite{ames2019control}.
\begin{theorem}
    \label{th1}
    If $K_\alpha$ satisfies $\forall k$: $p_k>0$ and the initial condition $\mathbf{x}_0 \in \mathcal{C}_k$, then $h(\mathbf{x})$ is a valid exponential CBF and $\mathcal{C}$ is forward invariant.
\end{theorem}
We refer the reader to \cite{ames2019control} for additional details on the proof of forward invariance and the linear control theory tools used for the choice of $K_\alpha$.

\section{Methodology}
\subsection{Problem Overview and Assumptions}
We consider a set of homogeneous quadrotors $q_i \in \mathcal{Q}$ and a set of obstacles $o_j \in \mathcal{O}$. The notation $e_j$ refers to an entity that can be either a quadrotor or obstacle, the index $i$ is used specifically for the ego quadrotor and $j$ for the other surrounding entities.
The state and control inputs of a quadrotor $q_i$ can be described as
\begin{equation*}
    \mathbf{x}_i = \begin{bmatrix}
    \mathbf{p}_i^\top & \mathbf{v}_i^\top & \mathbf{q}_i^\top & \boldsymbol{\omega}_i^\top
\end{bmatrix}^\top, \mathbf{u}_i=\begin{bmatrix}
    u_{i0} & u_{i1} & u_{i2} & u_{i3}
\end{bmatrix}^\top,
\end{equation*}
where $\mathbf{p}_i\in \mathbb{R}^3$ and $\mathbf{v}_i\in \mathbb{R}^3$ are respectively the position and linear velocity of the quadrotor in the inertial frame, $\mathbf{q}_i\in \mathbb{R}^4$ the rotation in quaternions from the quadrotor' body frame to the inertial frame, $\boldsymbol{\omega}_i\in \mathbb{R}^3$ the angular velocity in the body frame, and $\{u_{ik}\in \mathbb{R}, k\in[0,...,3]\}$ the motor thrusts of the quadrotor. The dynamic equations of $q_i$ are as presented in \cite{saviolo2023learning}. The objective is to optimize the control inputs $\mathbf{u}_i$ of each quadrotor $q_i$ in a decentralized fashion to track reference trajectories $\mathbf{x}_i^*$, while satisfying state, motor thrusts, and collision avoidance constraints.


In a decentralized setting, quadrotors can only rely on local communication and sensing capabilities to detect neighboring objects $e_j$, including quadrotors $q_j \in \mathcal{N}_i$ and obstacles $ o_j \in \mathcal{N}_{io}$. We denote $R_d$ and $R_{do}$ the quadrotors' detection ranges for other robots and obstacles, respectively. $R_d$ and $R_{do}$ can be different when quadrotors communicate with each other and perceive obstacles with local sensors. The terms $\mathcal{N}_i$ and  $\mathcal{N}_{io}$ are defined as follows
\begin{align*}
    \mathcal{N}_i &= \{q_j \in \mathcal{Q}, \quad \|\mathbf{p}_j - \mathbf{p}_i\| \leq R_d\}, \\ 
    \mathcal{N}_{io} &= \{o_j \in \mathcal{O}, \quad \|\mathbf{p}_j - \mathbf{p}_i\| \leq R_{do}\}.
\end{align*}
We assume that each quadrotor $q_i$ can estimate the relative position $\mathbf{p}_{rel,ij} = \mathbf{p}_j - \mathbf{p}_i$ and relative velocity $\mathbf{v}_{rel,ij} = \mathbf{v}_j - \mathbf{v}_i$ of their neighbors $e_j$. 

\subsection{Decentralized NMPC with ECBFs}
In the following, all variables are time-dependent, but we drop the time notation for better clarity. 
We formulate a nonlinear optimization problem over the prediction horizon $T$ for a quadrotor $q_i$ in continuous time
\begin{subequations}
\begin{align} 
    &\argmin_{\mathbf{u}_i \in \mathbb{R}^4}&& \int_{t_0}^{t_0 + T} \bigg( \|\mathbf{x}_i - \mathbf{x}_{i}^*\|_Q^2 + \|\mathbf{u}_i - \mathbf{u}_{i}^*\|_R^2 \bigg) \, dt \label{quad_cost}\\
   &\text{s.t.} \quad \forall t,  &&\dot{\mathbf{x}}_i = f(\mathbf{x}_i) + g(\mathbf{x}_i) \mathbf{u}_i, \label{dyn_cons}\\
   &&&\mathbf{x}_i \in \mathcal{X},~\mathbf{u}_i \in \mathcal{U}, \label{xu_cons}\\
   &&&G_{ij}(\mathbf{x}_i,\mathbf{x}_j,\mathbf{u}_i) \geq 0,~ \forall e_j \in \mathcal{N}_i \cup \mathcal{N}_{io} \label{h_cons},
\end{align}
\end{subequations}
where eq.~\eqref{quad_cost} represents the quadratic objective function minimizing the distance to the reference trajectory $\mathbf{x}_i^*$ and the reference control inputs $\mathbf{u}_i^*$, eq.~\eqref{dyn_cons} the dynamics of the quadrotor in a nonlinear control affine form, eq.~\eqref{xu_cons} the state and motor thrusts constraints, and eq.~\eqref{h_cons} the pair-wise ECBF constraints for collision avoidance between $q_i$ and each of its neighbors $e_j$. We elaborate the formulation of these ECBFs in the following.

\begin{figure}[t]
    \centering
    \includegraphics[width=0.7\columnwidth]{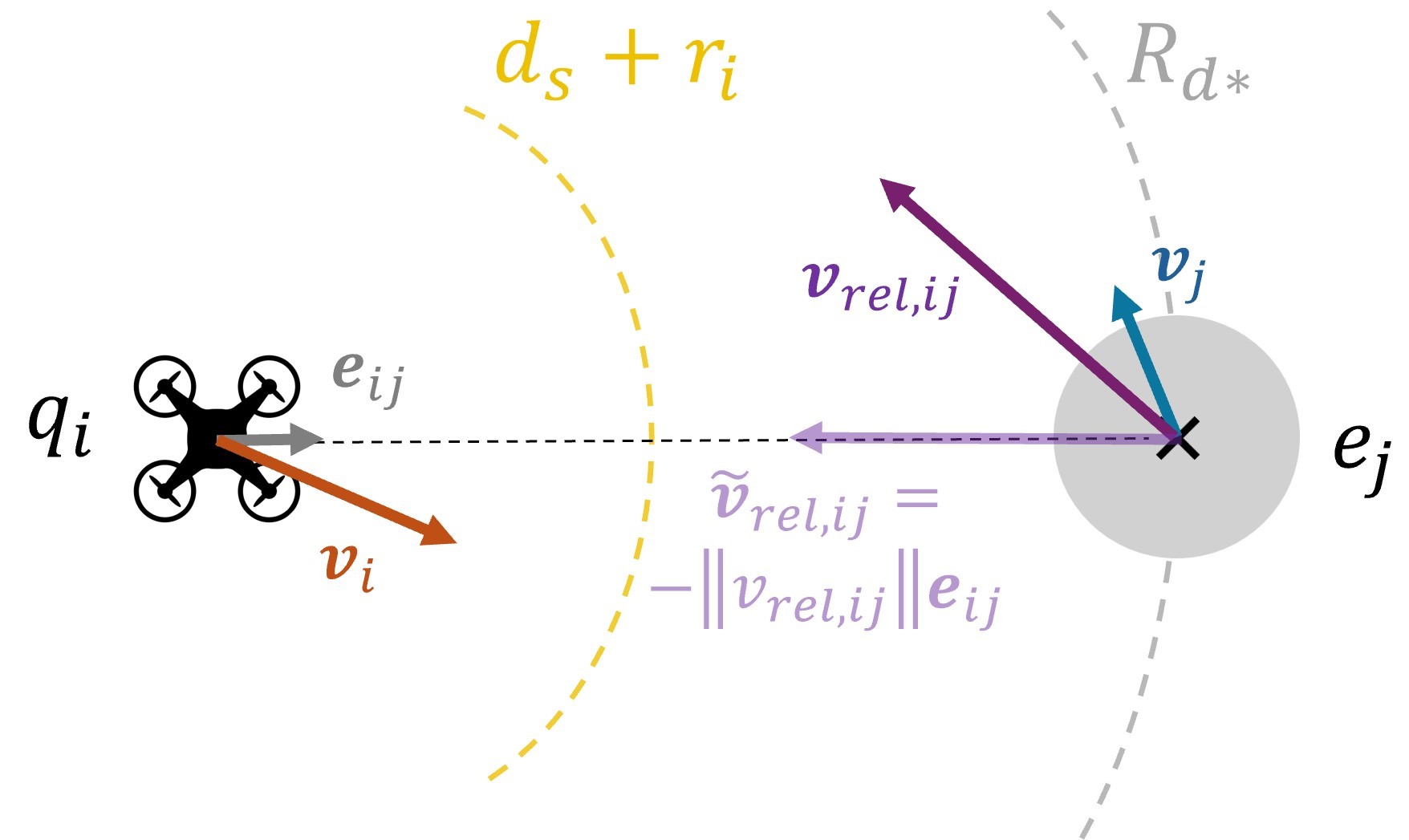}
    \vspace{-5pt}
    \caption{Relative velocity conservative approximation between quadrotor $q_i$ and obstacle/quadrotor $e_j$.}
    \label{fig:vrel_approx}
        \vspace{-10pt}
\end{figure}

We consider the quadrotors safe if they are at a certain distance $d_s$ from each other and from the obstacles. The safe set $\mathcal{C}_i$ for quadrotor $q_i$ is then
\begin{equation}
    \mathcal{C}_i = \hspace{-5pt}\inter_{e_j \in \mathcal{N}_i \cup \mathcal{N}_{io}}\hspace{-5pt} \mathcal{C}_{ij} = \hspace{-5pt}\inter_{e_j \in \mathcal{N}_i \cup \mathcal{N}_{io}} \hspace{-5pt}\{\mathbf{x}_i \in \mathcal{X}, h_{ij}(\mathbf{x}_i,\mathbf{x}_j) \geq 0\}, 
\end{equation}
with the pair-wise barrier function
\begin{equation}
    h_{ij}(\mathbf{x}_i,\mathbf{x}_j) = \|\mathbf{p}_{rel,ij}\|^2 - (d_s+ r_i + r_j)^2, 
\end{equation}
where $r_i$ and $r_j$ are the radial dimensions inferred by $q_i$ and $e_j$ respectively.
The barriers $h_{ij}(\mathbf{x}_i,\mathbf{x}_j)$ are ECBFs of relative-degree $r=2$ since the control inputs $\mathbf{u}_i = \begin{bmatrix}
    u_{i0} & u_{i1} & u_{i2} & u_{i3}
\end{bmatrix}^\top$ appear in the acceleration expression $\dot{\mathbf{v}}$~\cite{saviolo2023learning}. It follows that
\begin{align*}
    G_{ij}(\mathbf{x}_i, \mathbf{x}_j, \mathbf{u}_i) &= \ddot{h}_{ij} (\mathbf{x}_i, \mathbf{x}_j, \mathbf{u}_i) + \alpha_2 \dot{h}_{ij}(\mathbf{x}_i, \mathbf{x}_j) \\
   &+ \alpha_1 h_{ij}(\mathbf{x}_i, \mathbf{x}_j) \geq 0,
\end{align*}
where $\alpha_1$ and $\alpha_2$ need to be chosen according to the conditions presented in Theorem \ref{th1}.
To account for the unpredictability of the robots' motions, i.e. possible abrupt changes of direction toward $q_i$, we enforce safer predictions by considering the relative velocity $\mathbf{v}_{rel,ij}$ always pointing from $q_i$ to $e_j$ (see Fig.~\ref{fig:vrel_approx}). The result is a conservative approximation of the relative velocity, $\Tilde{\mathbf{v}}_{rel,ij}$
\begin{equation*}
    \Tilde{\mathbf{v}}_{rel,ij} =- \| \mathbf{v}_{rel,ij}\| \frac{\mathbf{p}_{rel,ij}}{\|\mathbf{p}_{rel,ij}\|} = -\| \mathbf{v}_{rel,ij}\| \mathbf{e}_{ij},
\end{equation*}
with $\mathbf{e}_{ij} = \frac{\mathbf{p}_{rel,ij}}{\|\mathbf{p}_{rel,ij}\|}$.
In that way, quadrotor $q_i$ always assumes the neighbor $e_j$ is coming towards it.
Moreover, during the NMPC prediction horizon $T$, quadrotor $q_i$ assumes that its neighbors will keep their velocity constant, resulting in
\begin{equation*}
    \forall e_j \in \mathcal{N}_i \cup \mathcal{N}_{io}, ~\forall t \in [t_0,t_0 + T],~\dot{\mathbf{v}}_j (t) =0.
\end{equation*}

These assumptions yield the following expressions of the time-derivatives of $h_{ij}$
\begin{align*}
    & \dot{h}_{ij}(\mathbf{x}_i,\mathbf{x}_j) = 2 \hspace{2pt}\mathbf{p}_{rel,ij} \cdot \Tilde{\mathbf{v}}_{rel,ij}, \\
    & \ddot{h}_{ij}(\mathbf{x}_i,\mathbf{x}_j, \mathbf{u}_i)  = 2\hspace{2pt} \|\Tilde{\mathbf{v}}_{rel,ij}\|^2 - 2 \hspace{2pt}\mathbf{p}_{rel,ij} \cdot \mathbf{\dot{v}}_i.
\end{align*}

\subsection{Safety Guarantees Analysis with Limited Detection Range}
The objective of this theoretical analysis is to calculate the minimum detection range possible, given $\alpha_1$ and $\alpha_2$, that will not compromise the safety guarantees of the ECBFs, i.e. the feasibility of the safety constraints and the foward invariance of the safe set. 
We assume that the ECBFs' gains $\alpha_1$ and $\alpha_2$ are pre-tuned according to the conditions of Theorem \ref{th1} and without any detection restriction.
\vspace{5pt}
\subsubsection{Minimum detection range for a pair $(q_i,e_j)$} \hfill
\vspace{5pt}

First, let us consider a quadrotor $q_i$ that detects a neighbor $e_j$ at a distance $R_{d*}$ at some time $t_0$ and activates its safety constraint $G_{ij}$. $R_{d*}$ corresponds to a quadrotor-to-quadrotor detection range $R_{d}$ or a quadrotor-to-obstacle detection range $R_{do}$. Hence, the relative position between $q_i$ and $e_j$ is $\mathbf{p}_{rel,ij}(t_0) = R_{d*} \mathbf{e}_{ij}$. 

\begin{definition}[Conservative bound $\hat{R}_{d*}$] \label{def_con}
The conservative bound $\hat{R}_{d*}$ is the minimum value of $R_{d*}$ so that, the following conditions are satisfied at $t_0$:
\begin{enumerate}[label=(\roman*)]
    \item $\boundinf\limits_{\mathbf{u}_i\in \mathcal{U}} \boundinf\limits_{\Tilde{\mathbf{v}}_{rel,ij}} G_{ij}(R_{d*}, \Tilde{\mathbf{v}}_{rel,ij}, \mathbf{u}_i) \geq 0$, \label{item1bis}
    \item $\boundinf\limits_{\Tilde{\mathbf{v}}_{rel,ij}} \dot{h}_{ij}(R_{d*}, \Tilde{\mathbf{v}}_{rel,ij})+p_1 h_{ij}(R_{d*})\geq 0$, \label{item2bis}
    \item $h_{ij}(R_{d*})\geq 0$ \label{item3bis}
\end{enumerate}
These conditions guarantee $h_{ij}$ stays a valid ECBF and $\mathcal{C}_{ij}$ forward invariant whatever the relative velocity $\Tilde{\mathbf{v}}_{rel,ij}$ and action $\mathbf{u}_i$ of quadrotor $i$ at $t_0$. Condition \ref{item1bis} enforces the satisfaction of the safety constraint $G_{ij}$. Conditions \ref{item2bis} and \ref{item3bis} ensures that the initial conditions stated in Theorem \ref{th1} are not compromised.
\end{definition}

{\begin{proposition}
\label{prop_cons}
    $\hat{R}_{d*}$ exists and is equal to
    \begin{equation*}
        \hat{R}_{d*} = \min \biggl\{ \frac{-\hat{b}_{(i)}+\sqrt{\hat{\Delta}_{(i)}}}{2 \hat{a}_{(i)}}, \frac{-\hat{b}_{(ii)}+\sqrt{\hat{\Delta}_{(ii)}}}{2 \hat{a}_{(ii)}}, d_s+r_i+r_j  \biggr\}
    \end{equation*} where \vspace{-5pt}
    \begin{align*}
        &\hat{a}_{(i)} = \alpha_1, && \hat{a}_{(ii)} = p_1, \\
        & \hat{b}_{(i)} = -2(a_{max,i}+\alpha_2v_{rel,max}), && \hat{b}_{(ii)} = -2v_{rel,max}, \\
        & \hat{c}_{(i)} = 2v_{rel,max}^2 \!- \alpha_1(d_s \!+\!r_i\!+\!r_j)^2, && \hat{c}_{(ii)} = p_1(d_s\!+\!r_i\!+\!r_j)^2, \\
        & \hat{\Delta}_{(i)} = \hat{b}_{(i)}^2 \!- \!4\hat{a}_{(i)}\hat{c}_{(i)}, &&
        \hat{\Delta}_{(ii)} = \hat{b}_{(ii)}^2 \!-\! 4\hat{a}_{(ii)}\hat{c}_{(ii)}
    \end{align*}
\end{proposition}}
\noindent\textit{\textbf{Proof \ref{prop_cons}:}} see Appendix \ref{proof1_appendix}.

We would like to highlight that the bound $\hat{R}_{d*}$ derived above is very conservative, because it guarantees the safety constraint satisfaction between $q_i$ and $e_j$ for all possible actions $\mathbf{u}_i$ quadrotor $i$ takes at $t_0$. In the following, we propose a less conservative bound $\check{R}_{d*}$ that guarantees the existence of at least one control input $\mathbf{u}_i \in \mathcal{U}$ that can satisfy the safety constraint and preserves the forward invariance of the safe set $\mathcal{C}_{ij}$.

\begin{definition}[Non-conservative bound $\check{R}_{d*}$] \label{def_ncon}
The non-conservative bound $\hat{R}_{d*}$ is the minimum value of $R_{d*}$ so that the following conditions are satisfied at $t_0$:
\begin{enumerate}[label=(\roman*)]
    \item $\boundsup\limits_{\mathbf{u}_i \in \mathcal{U}} \boundinf\limits_{\Tilde{\mathbf{v}}_{rel,ij}} G_{ij}(R_{d*}, \Tilde{\mathbf{v}}_{rel,ij}, \mathbf{u}_i) \geq 0$, \label{item1}
    \item $\boundinf\limits_{\Tilde{\mathbf{v}}_{rel,ij}} \dot{h}_{ij}(R_{d*}, \Tilde{\mathbf{v}}_{rel,ij})+p_1 h_{ij}(R_{d*})\geq 0$, \label{item2}
    \item $h_{ij}(R_{d*})\geq 0$ \label{item3}
\end{enumerate}
Based on Theorem \ref{th1} and Definition \ref{def_ecbf}, these conditions guarantee $h_{ij}$ stays a valid ECBF and $\mathcal{C}_{ij}$ forward invariant whatever the relative velocity $\mathbf{v}_{rel,ij}$ at $t_0$.
\end{definition}

{
\begin{proposition}
\label{prop_ncons}
    $\check{R}_{d*}$ exists and is equal to
    \begin{equation*}
        \check{R}_{d*} = \min \biggl\{ \frac{-\check{b}_{(i)}+\sqrt{\check{\Delta}_{(i)}}}{2 \check{a}_{(i)}}, \frac{-\check{b}_{(ii)}+\sqrt{\check{\Delta}_{(ii)}}}{2 \check{a}_{(ii)}}, d_s+r_i+r_j  \biggr\}
    \end{equation*} where \vspace{-5pt}
    \begin{align*}
        &\check{a}_{(i)} = \alpha_1, && \check{a}_{(ii)} = p_1, \\
        & \check{b}_{(i)} = 2(a_{max,i}-\alpha_2v_{rel,max}), && \check{b}_{(ii)} = -2v_{rel,max}, \\
        & \check{c}_{(i)} = 2v_{rel,max}^2 \!- \alpha_1(d_s \!+\!r_i\!+\!r_j)^2, && \check{c}_{(ii)} = p_1(d_s\!+\!r_i\!+\!r_j)^2, \\
        & \check{\Delta}_{(i)} = \check{b}_{(i)}^2 \!- \!4\check{a}_{(i)}\check{c}_{(i)}, &&
        \check{\Delta}_{(ii)} = \check{b}_{(ii)}^2 \!-\! 4\check{a}_{(ii)}\check{c}_{(ii)}
    \end{align*}
\end{proposition} 
}
\noindent\textit{\textbf{Proof \ref{prop_ncons}:}} see Appendix \ref{proof2_appendix}.

In the next section, we analyze the compatibility of multiple ECBF constraints when the quadrotors are subject to these minimal detection ranges.
\vspace{5pt}
\subsubsection{Compatibility of multiple ECBF constraints} \hfill
\begin{proposition}[Compatibility under the conservative bound $\hat{R}_{d*}$]
\label{prop_comp}
    If all neighbors $e_j$ are at a distance $R_{d*} \geq \hat{R}_{d*}$ from $q_i$ at $t_0$ when quadrotor $i$ activates its ECBF constraints, then the safe set of $q_i$ $\mathcal{C}_i = \inter_{e_j \in \mathcal{N}_i \cup \mathcal{N}_{io}}\hspace{-5pt} \mathcal{C}_{ij}$ is forward invariant.
\end{proposition}
\noindent\textbf{\textit{Proof \ref{prop_comp}:}} see Appendix \ref{proof3_appendix}.

As a result, under the limited detection range $\hat{R}_{d*}$, the safe set of the quadrotor team $\mathcal{C}_t = \inter \mathcal{C}_i$ is foward invariant with the gains $\alpha_1$ and  $\alpha_2$.

These guarantees do not hold anymore for the non-conservative bound $\check{R}_{d*}$. Indeed, an action $\mathbf{u}_i$ of $q_i$ towards a neighbor $e_j$ that satisfies condition \ref{item1} of Definition \ref{def_ncon}, does not guarantee the satisfaction of condition~\ref{item1} for all other neighbors. 
In the next section, we demonstrate the effectiveness of the non-conservative bound to keep the quadrotor team safe in practice, through extensive experiments.

\section{Results}
We deploy our control framework in simulation and real-world experiments. For each experiment, the quadrotors need to reach their goal objectives while navigating through static obstacles and avoiding their teammates. We do not simulate moving obstacles since they can be treated as quadrotors given our decentralized design and assumptions. The quadrotors access to each other's relative information through communication. They track a reference trajectory which is a minimum jerk line from their initial position to their goal position.

In the previous section, we derived the theoretical bounds for the minimum detection range required to preserve the safety guarantees in continuous time. In practice, to account for the discretization of the NMPC solver, we discretize these bounds with a one-step Euler integration:
\begin{align*}
    & \hat{R}_{dd*} = \hat{R}_{d*} + dt v_{rel,max}, 
    & \check{R}_{dd*} = \check{R}_{d*} + dt v_{rel,max},
\end{align*}
where $dt$ is the discretization time step and $dt v_{rel,max}$ corresponds to the maximum possible displacement of two objects (robots or obstacles) towards each other during $dt$. 

For all the experiments, the radius of the robots is $r_q =0.2~\si{m}$ and their acceleration limit has been evaluated at $a_{max}= 2~\si{m/s^{2}}$ according to our quadrotors' hardware characteristics including mass, maximum motors' rpm and time of response. We decide to enforce a safety distance of $d_s = 0.4~\si{m}$ among quadrotors, and $d_{so} = 0.2~\si{m}$ between quadrotors and obstacles. We use acados \cite{verschueren2022acados}, SQP-RTI, a horizon of $T = 1~\si{s}$ and $dt = 0.1~\si{s}$ to solve our NMPC. The robots communicate using ROS2\footnote{\url{https://www.ros.org/}}. The ECBF gains $\alpha_1 = 36 $ and $\alpha_2 = 22$ are pre-tuned without any detection restriction so that they satisfy the conditions stated in Theorem \ref{th1} in continuous time and in discrete time simulation.
\begin{figure}[t]
    \centering
    \includegraphics[width=0.94\columnwidth]{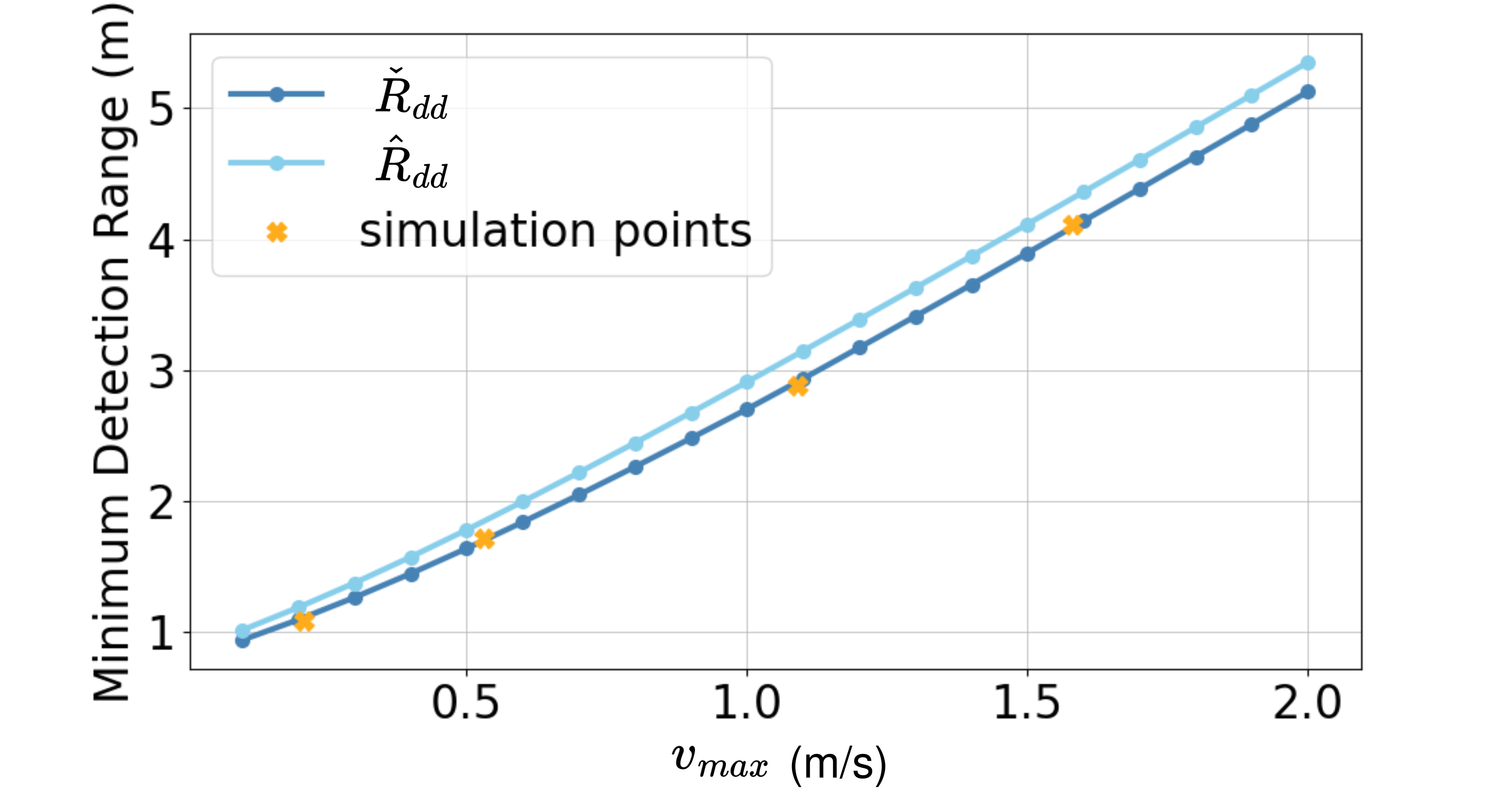}
    \vspace{-4.5pt}
    \caption{Minimum theoretical and simulated detection ranges between two quadrotors in discrete time when switching positions at different velocities and with a maximum acceleration of $2~\si{m/s^{2}}$.}
    \label{fig:rd_analysis}
    \vspace{-10pt}
\end{figure}

First, we validate the values of the detection range bounds in simulation by testing two quadrotors switching positions and flying at different velocities $v_{max}$, according to the worst-case scenario previously presented. We reduce the detection range until the safety distance is violated and compare the limit found with the theoretical discrete-time bounds in Fig. \ref{fig:rd_analysis}. As expected, the simulation data points match with the non-conservative lower bound.

In the rest of the experiments, we test the efficacy of the non-conservative bounds for quadrotor-to-quadrotor detection $\check{R}_{dd}$ and quadrotor-to-obstacle detection $\check{R}_{ddo}$, derivated for homogeneous robots and static obstacles, $
    \check{R}_{dd} = \check{R}_{dd} + 2dt v_{max},~\text{and}~ \check{R}_{ddo} = \check{R}_{do} + dt v_{max}.$
The maximal velocity of the robots is set at $v_{max} = 1.5~\si{m/s}$. We refer the reader to the attached multimedia material for additional illustrations and videos of the experiments.

\subsection{Simulation Results}
\begin{figure}[t]
    \centering
    \includegraphics[width=\columnwidth]{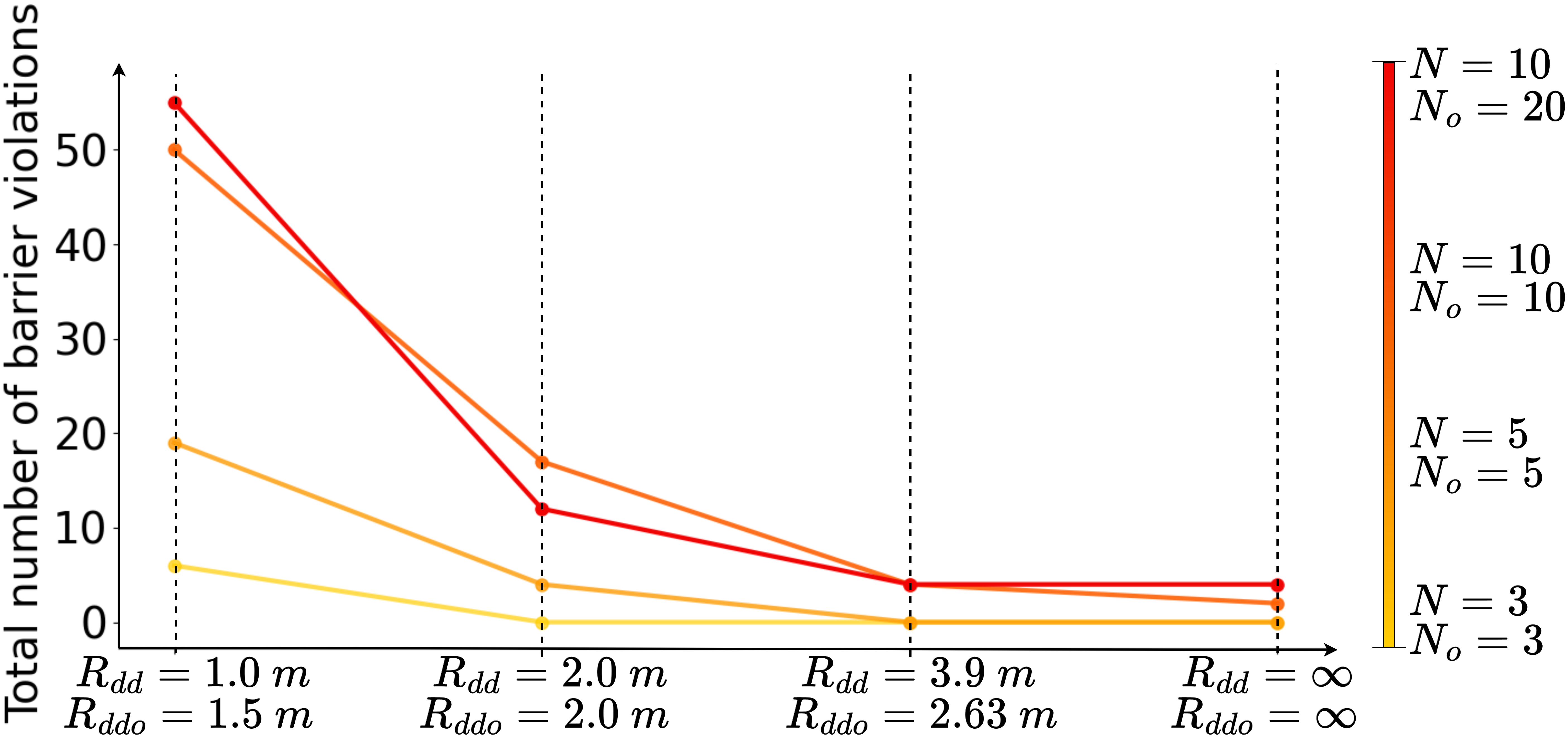}
    \caption{Total number of barrier violations in simulations as a function of the number of robots $N$, number of obstacles $N_o$, and detection range restrictions $R_{dd}$ between quadrotors and $R_{ddo}$ between quadrotors and obstacles.}
    \label{fig:sim_results}
    \vspace{-10pt}
\end{figure}

We evaluate the collision avoidance performance for simulated environments of increasing complexity, characterized by the number of robots $N$, the number of obstacles $N_o$, and the detection ranges $R_{dd}$ and $R_{ddo}$. We consider $10$ simulation runs for each setup $(N,N_o)$, where initial, goal and obstacle positions are uniformly randomized in the environment bounds $[x,y,z] \in [-8,8]\!\times\![-8,8]\!\times\![0.5,2]$. The obstacles' radii $r_o$ are also randomized and range from $0.1$ to $1~\si{m}$. The detection range bounds $\check{R}_{dd*}$ are $
    \check{R}_{dd} = 3.60+ 2dtv_{max} = 3.9~\si{m},~\check{R}_{ddo} = 2.48+ dtv_{max} = 2.63~\si{m}$,
where $\check{R}_{ddo}$ is determined for obstacles of maximum dimension of $1~\si{m}$. Various detection range scenarios are tested: no restriction, i.e. $R_{dd} = R_{ddo} = \infty$, the bounds $\check{R}_{ddo}$ and $\check{R}_{ddo}$, lower ranges $R_{dd} = 2.0~\si{m}$ and $R_{ddo} = 2.0~\si{m}$, and very restrictive ranges $R_{dd} = 1.0~\si{m}$ and $R_{ddo} = 1.5~\si{m}$ just above the safety distances $d_s+2r_q$ and $d_s+r_q+\max(r_o)$. Figure \ref{fig:sim_results} reports the total number of barrier violations across the 10 realizations of each setup. A barrier violation happens when a robot violates the safety distance with another robot or an obstacle. As expected, the more complex is the scenario, i.e. the higher the density of robots and obstacles in the environement and the lowest the detection range, the more the safety is compromised. Specifically, for $R_{dd} = 1.0~\si{m}$ and $R_{ddo} = 1.5~\si{m}$, the number of barrier violations increases dramatically, indicating that overly restrictive detection ranges can severely compromise the safety of the robots.

However, the proposed lower bound, $\check{R}_{dd*}$, does not affect the overall safety of the system when compared to an infinite detection range, demonstrating the practical effectiveness of the bound. Furthermore, with this detection range, all robots effectively stay safe for $N \leq 5$ and $N_o \leq 5$. In more dense environments, with $N = 10$ and $N_o \geq 10$, the total number of violations remains relatively low ($\leq 4$). This highlights the strong performance of our control approach. The number of violations in scenarios with unlimited detection range is not always zero. This occurs because, aside from detection limitations, the NMPC can be infeasible in deadlock situations typical to dense environments and decentralized settings. This is a limitation of our control approach. 

\begin{figure*}[t]
    \centering
    \includegraphics[width=2\columnwidth]{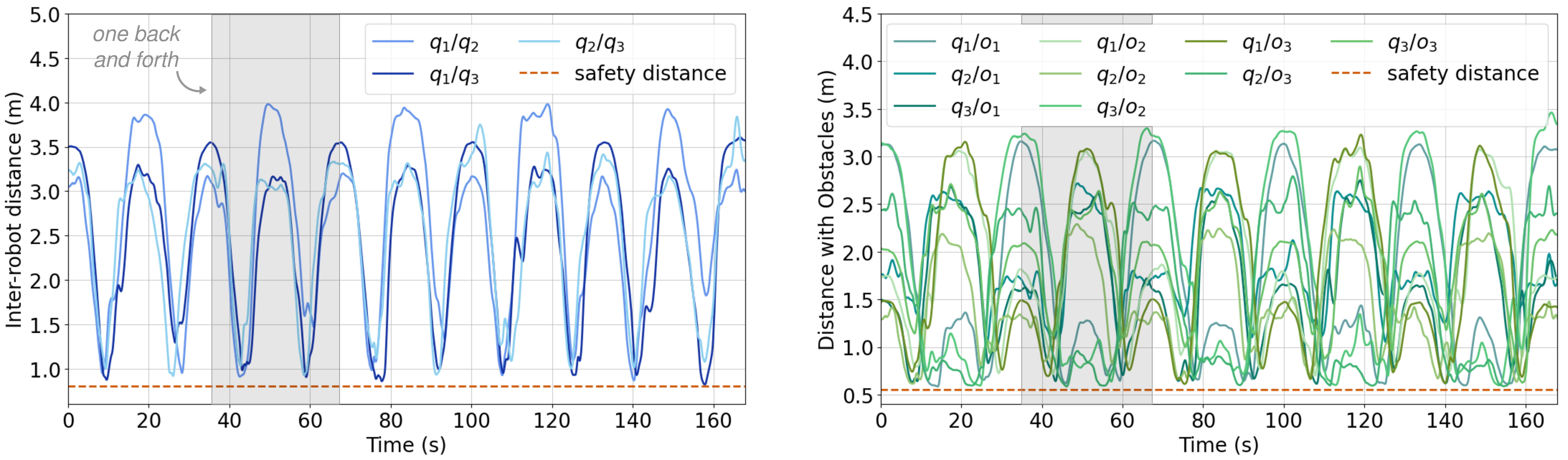}
    \caption{Distances between quadrotors $q_i$ and obstacles $o_j$ over time. The grey zone denotes one back and forth motion.}
    \label{fig:rw-exp-results}
    \vspace{-10pt}
\end{figure*}

\subsection{Real-world Experiments}
We demonstrate the practical applicability of our solution in a challenging scenario illustrated in Fig.~\ref{fig:rw_exp}, where three quadrotors need to cross each other while constrained by three static obstacles of radius $0.15~\si{m}$. The robots navigate in an indoor testbed of $10\times6\times4~\si{m^3}$ with a Vicon
motion capture system for localization. We employ $3$ custom built quadrotors equipped with a Qualcomm\textsuperscript{\textregistered} $\text{Snapdragon}^{\text{TM}}$ VOXL\textsuperscript{\textregistered}2
board~\cite{LoiannoRAL2017}. 
We constraint the robots to the non-conservative detection range bounds
$\check{R}_{dd} = 3.60+ 2dtv_{max} = 3.9~\si{m},~\text{and}~\check{R}_{ddo} = 1.7+ dtv_{max} = 1.85~\si{m}$.
The NMPC runs on-board on each quadrotor, at a frequency of $\sim 160~\si{Hz}$. The robots go back and forth from their initial positions to their goal positions, illustrated in Fig.\ref{fig:rw_exp}, $5$ times within the same experiment. Because of real-world uncertainties related to noisy velocity measurements (this is also the case for vicon since the velocity is obtained as the numerical derivative of the position) or communication delays, their trajectories are different during each travel. Fig. \ref{fig:rw-exp-results} reports the inter-robot and robot-to-obstacle distances over time, and the corresponding safety distances $d_s + 2 r_q = 0.8~m$ and $d_{so} + r_q+r_o = 0.55~m$ enforced. As illustrated, the safety constraints are  successfully  satisfied for each robot during the repeated $5$ back and forth motions, which validates the efficacy of our safe controller and the non-conservative detection range bound proposed.
\section{Conclusion}
In this work, we introduced a novel decentralized NMPC with ECBF-based safety constraints to control a team of quadrotors, under motor thrust constraints and limited sensing and communication capabilities. We examined how a restricted detection range affects ECBF safety guarantees, calculating theoretical lower bounds to ensure robots' safety and validating them in simulations of increasing complexity. We also tested our solution on real quadrotors and proved its efficacy within the derived detection range. However, the NMPC may encounter feasibility issues in crowded environments. 

Future research will focus on developing deadlock handling algorithms to improve its robustness. Moreover, to eliminate the need of pre-tuned ECBFs, adaptive gains will be explored for enhanced flexibility and resilience in the presence of limited detection but also information uncertainties or communication delays.

\section{Appendix}
\subsection{Proof of Proposition~\ref{prop_cons}} \label{proof1_appendix}
The infimum $\boundinf\limits_{\Tilde{\mathbf{v}}_{rel,ij}} G_{ij}$ is found for the worst case scenario when $q_i$ and $e_j$ are coming towards each other with their maximal possible velocities $v_{max,i}$ and $v_{max,j}$. The relative velocity is then
\vspace{-5pt}
{
\begin{align*}
    &\Tilde{\mathbf{v}}_{rel,ij} = -(v_{max,j}+v_{max,i}) \mathbf{e}_{ij} =-v_{rel,max}\mathbf{e}_{ij}.
\end{align*}
Substituting the worst-case $\Tilde{\mathbf{v}}_{rel,ij}$ and $\mathbf{p}_{rel,ij} = R_{d*}$ in the barrier function equations, we get
\begin{align*}
    &h_{ij}(R_{d*}) = R_{d*}^2 - (d_s+ r_i + r_j)^2, \\ &\boundinf\limits_{\Tilde{\mathbf{v}}_{rel,ij}} \dot{h}_{ij}(R_{d*}, \Tilde{\mathbf{v}}_{rel,ij}) = - 2 R_{d*} v_{rel,max}, \\
    & \boundinf\limits_{\Tilde{\mathbf{v}}_{rel,ij}} \ddot{h}_{ij}(R_{d*}, \Tilde{\mathbf{v}}_{rel,ij},\mathbf{u}_i) = 2 v_{rel,max}^2-2(R_{d*} \mathbf{e}_{ij})\cdot \dot{\mathbf{v}}_i.
\end{align*}}
The infimum $\boundinf\limits_{\mathbf{u}_i\in \mathcal{U}} G_{ij}$ is found when $q_i$ has the maximum acceleration possible $a_{max,i}$ towards $e_j$:
{
\begin{align*}
\boundinf\limits_{\mathbf{u}_i \in \mathcal{U}} -2(R_{d*} \mathbf{e}_{ij})\cdot \dot{\mathbf{v}}_i = -2R_{d*} a_{max,i}.
\end{align*}}
Substituting these expressions in the conditions \ref{item1bis}, \ref{item2bis} and \ref{item3bis} from Definition \ref{def_con}, we obtain the following inequalities
\begin{enumerate}[label=(\roman*),left=40pt]
     \item $\hat{a}_{(i)} R_{d*}^2 + \hat{b}_{(i)} R_{d*} + \hat{c}_{(i)}  \geq 0$,
    \item $\hat{a}_{(ii)} R_{d*}^2 + \hat{b}_{(ii)} R_{d*} + \hat{c}_{(ii)}  \geq 0$,
    \item $R_{d*}^2 - (d_s+r_i+r_j)^2\geq 0$. 
\end{enumerate}
 Each polynomial ($k$) has two roots $R_{1k}$ and $R_{2k}$, which results in the solutions of the inequality ($k$)
 \begin{equation*}
     R_{d*} \in (-\infty, R_{1k}]\,\,\,\text{and}\,\,\, R_{d*} \in [R_{2k},+\infty).
 \end{equation*} 
 We retain $R_{2k} = \max\{R_{1k},R_{2k}\}$ as the only physically possible solution, since $\hat{R}_{d*}$ is positive and $e_j$ is at a distance greater than $R_{d*}$ from $q_i$ at $t \leq t_0$. Then, in order to satisfy all three conditions, the following expression of $\hat{R}_{d*}$ holds:
 $\hat{R}_{d*} = \min \{R_{2i},R_{2ii},R_{2iii} \}$. The values of $R_{2k}$ are as stated in Proposition \ref{prop_cons}, according to the well-known expression of the roots of a general second-order polynomial.

\subsection{Proof of Proposition \ref{prop_ncons}} \label{proof2_appendix}

As previoulsy, the infimum bounds are found at the worst-case scenario. The supremum $\boundsup\limits_{\mathbf{u}_i \in \mathcal{U}} G_{ij}$ is found when $q_i$ takes the control input $\mathbf{u}_i \in \mathcal{U}$ that results in the maximum possible deceleration away from $e_j$:
\begin{align*}
\boundsup\limits_{\mathbf{u}_i \in \mathcal{U}} -2(R_{d*} \mathbf{e}_{ij})\cdot \dot{\mathbf{v}}_i = 2R_{d*} a_{max,i}.
\end{align*}
Then, a similar proof than Proof \ref{prop_cons} can be derived where the three conditions \ref{item1}, \ref{item2} and \ref{item3} are equivalent to second-order polynomial inequalities, and $\check{R}_{d*}$ is the minimum of the corresponding second roots.

\subsection{Proof of Proposition~\ref{prop_comp}} \label{proof3_appendix}

If a neighbor $e_j$ is at a distance $R_{d*} \geq \hat{R}_{d*}$ from $q_i$ at $t_0$, then $\mathcal{C}_{ij}$ is forward invariant according to Definition \ref{def_con} and Proposition \ref{prop_cons}. Since this is guaranteed whatever the values of $\mathbf{v}_{rel,ij}$ and $\mathbf{u}_i$, the conditions of Definition \ref{def_con} are compatible across all neighbors $e_j$. So, if all neighbors $e_j$ are at $R_{d*} \geq \hat{R}_{d*}$ at $t_0$, then all $\mathcal{C}_{ij}$ are forward invariant and so is $\mathcal{C}_i$.


\bibliographystyle{IEEEtran}
\bibliography{IEEEabrv,references}

\end{document}